\documentclass[letterpaper, 10 pt, conference]{ieeeconf}
\IEEEoverridecommandlockouts    
\usepackage{graphics}           
\usepackage{times}              
\usepackage{amsmath}            
\usepackage{amssymb}            
\usepackage{graphicx}
\usepackage{algorithm}
\usepackage[noend]{algpseudocode}
\usepackage{booktabs}
\usepackage{color}
\definecolor{instructioncolor}{rgb}{.5,.5,.5}

\usepackage[font=small]{caption}

\def\secref#1{Sec.~\ref{#1}}
\def\figref#1{Fig.~\ref{#1}}
\def\tabref#1{Tab.~\ref{#1}}
\def\eqref#1{Eq.~(\ref{#1})}

\makeatletter
\usepackage{xspace}
\DeclareRobustCommand\onedot{\futurelet\@let@token\@onedot}
\def\@onedot{\ifx\@let@token.\else.\null\fi\xspace}

\def\etal{{et al}\onedot}
\makeatother

\def\etalcite#1{\etal~\cite{#1}}

\usepackage{array}
\newcolumntype{L}[1]{>{\raggedright\let\newline\\\arraybackslash\hspace{0pt}}m{#1}}
\newcolumntype{C}[1]{>{\centering\let\newline\\\arraybackslash\hspace{0pt}}m{#1}}
\newcolumntype{R}[1]{>{\raggedleft\let\newline\\\arraybackslash\hspace{0pt}}m{#1}}

\newcommand{\norm}[1]{\lVert#1\lVert}

\usepackage{hyperref}

\title{\LARGE \bf Long-Term Localization using Semantic Cues in Floor Plan Maps}

\author{Nicky Zimmerman \and Tiziano Guadagnino \and Xieyuanli Chen \and Jens Behley \and Cyrill Stachniss
  \thanks{All authors are with the University of Bonn, Germany. Cyrill Stachniss is additionally with the Department of Engineering Science at the University of Oxford, UK.}%
  \thanks{This work has partially been funded by the European Union’s Horizon 2020 research and innovation programme under grant agreement No~101017008~(Harmony).
  }%
}

\begin{document}
\maketitle
\thispagestyle{empty}
\pagestyle{empty}

\begin{abstract}
  %

  Lifelong localization in a given map is an essential capability for autonomous service robots. 
  In this paper, we consider the task of long-term localization in a changing indoor environment given sparse CAD floor plans. The commonly used pre-built maps from the robot sensors may increase the cost and time of deployment. Furthermore, their detailed nature requires that they are updated when significant changes occur. We address the difficulty of localization when the correspondence between the map and the observations is low due to the sparsity of the CAD map and the changing environment. To overcome both challenges, we propose to exploit semantic cues that are commonly present in human-oriented spaces. These semantic cues can be detected using RGB cameras by utilizing object detection, and are matched against an easy-to-update, abstract semantic map. The semantic information is integrated into a Monte Carlo localization framework using a particle filter that operates on 2D LiDAR scans and camera data. 
  We provide a long-term localization solution and a semantic map format, for environments that undergo changes to their interior structure and detailed geometric maps are not available. We evaluate our localization framework on multiple challenging indoor scenarios in an office environment, taken weeks apart. The experiments suggest that our approach is robust to structural changes and can run on an onboard computer. 
  We released the open source implementation\footnote{https://github.com/PRBonn/hsmcl} of our approach written in C++ together with a ROS wrapper.

\end{abstract}

\section{Introduction}
\label{sec:intro}

\begin{figure}[t]
  \centering
  \includegraphics[width=0.49\linewidth]{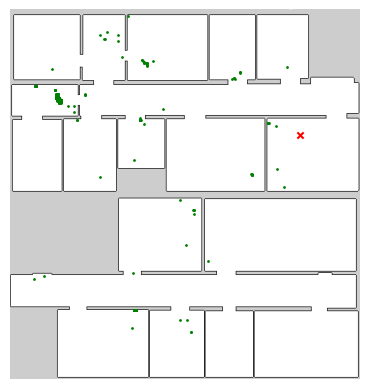}
  \includegraphics[width=0.49\linewidth]{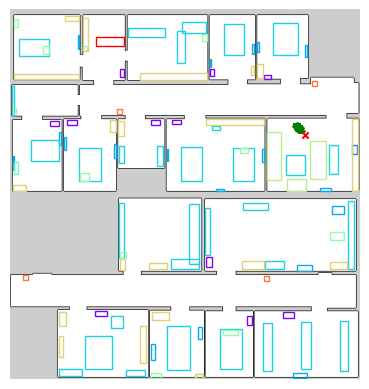}
  \caption{Floor plan maps include high degree of symmetry and low similarity to actual LiDAR measurements. This leads to multiple hypotheses that cannot be resolved correctly. We propose integrating semantic cues from a high level, abstract semantic map to assist with global localization. The red cross indicates the ground truth pose and the green dots are the particles. Left: 2D LiDAR MCL with multiple hypotheses. Right: Convergence to a single hypothesis when exploiting semantic cues, in an abstract semantic maps including various objects (colored rectangles). }
  \label{fig:motivation2}
\end{figure}

To operate autonomously in indoor environments, such as factories or offices, mobile robots must be able to determine their pose. For localization in a given map, there are two challenges: the changing nature of human-occupied environment and the quality of available maps. Precise, highly-detailed maps are an accurate representation of the environment only at the time they were captured, and they become outdated in the presence of ``quasi-static'' changes such as moving furniture, clutter, opening and closing doors. We describe ``quasi-static'' changes as long-lasting alterations (hours, days, weeks) that cause deviation between sensor observations and the given map, in contrast to dynamics such as humans and fast-moving objects.
The availability of feature-rich, dense maps is not guaranteed and construction of such maps can be costly. Therefore, autonomous robots benefit from localizing in sparse maps such as floor plans or hand-crafted room layouts as they are seldom affected by changes. Architectural drawings are familiar to inexpert users and can be easily updated with CAD software. As they capture persistent structures, they typically do not require updates. However, using these sparse maps is challenging due to the paramount discrepancies between the robot's observations of the environment and the information depicted in the maps. Additionally, floor plans lack geometric information necessary to localize in a highly repetitive indoor environment, as can be seen in \figref{fig:motivation2}.


 Additional sources of information can be used to overcome the challenges of global localization, and such cues have been frequently used by researchers to improve robot localization. For example, WiFi, an extremely prevalent utility, can aid in pose estimation by considering the signal strength~\cite{ito2014icra}. Textual information, constantly used by humans to navigate, is readily available in human-occupied environments. However, very few works consider textual cues for localization~\cite{cui2021iros}\cite{radwan2016icra}\cite{zimmerman2022iros}. 

 Another avenue is exploiting semantic information. The last decade was marked by significant advances in object detection~\cite{bochkovskiy2020arxiv}\cite{zhang2022arxiv} and semantic segmentation~\cite{he2017iccv}\cite{seichter2021icra}, where semantic cues can be efficiently inferred from images (with some fine-tuning). 
 The most commonly used map representation for robotics is an occupancy grid map~\cite{moravec1989sdsr}. However, human environments tend to be object-centric, and humans do not require precise metric information in order to navigate them~\cite{mendez2018sedar}\cite{yi2009iros}. Rather, humans rely on a small number of specific landmarks, and associate places with the objects present there. For this reason, we consider localization in a sparse, approximate map, that does not require an elaborate map acquisition process. No work on semantic localization in sparse maps with abstract and hierarchical semantic information exists to our knowledge.




The main contributions of this paper is a global localization system in floor plan maps that integrates semantic cues. We propose to leverage semantic cues to break the symmetry and distinguish between locations that appear similar or identical in the nondescript maps. Semantic information is commonly available in the form of furniture, machinery and textual cues and can be used to distinguish between spaces with similar layout. To avoid the complexity of building a 3D map from scans and to enable easy updates to semantic information, we present a 2D, high level semantic map. Thus, we present a format for abstract semantic map with an editing application and a sensor model for semantic information that complements LiDAR-based observation models. Additionally, we provide a way to incorporate hierarchical semantic information.  
Unlike most modern semantic-based SLAM approaches~\cite{chen2019iros}\cite{mccormac20183dv}\cite{runz2018ismar}\cite{wada2020cvpr}\cite{binbin2019icra}, our approach does not require a GPU and can run online on an onboard computer. 
In our experiments, we show that our approach is able to:
(i) localize in sparse floor plan-like map with high symmetry using semantic cues,
(ii) localize long-term without updating the map,
(iii) localize in previously unseen environment. 
(iv) localize the robot online using an onboard computer. These claims are backed up by the paper and our experimental evaluation.

\section{Related Work}
\label{sec:related}



Localization in 2D maps has been thoroughly researched~\cite{cadena2016tro}\cite{thrun2005probrobbook}\cite{trahanias2005ram}\cite{zafari2019cst}. Among the most robust and commonly-used approaches, are the probabilistic methods for pose estimation, including Markov localization by Fox \etalcite{fox1999jair}, the extended Kalman filter~(EKF) \cite{leonard1991tra} and particle filters, also known as Monte Carlo localization~(MCL) by Dellaert \etalcite{dellaert1999icra}. These works laid the foundation for localization using range sensors and cameras. 

Localization in detailed, feature-rich maps, usually constructed by range sensors, is extensively-studied~ \cite{moravec1985icra}, but few works address the problem of localization in sparse, floor plan-like maps, despite their benefits. Floor plans are readily-available in many facilities, and therefore do not depend on prior mapping. As they only include information on permanent structures, they do not require frequent updates when objects, such as furniture, are relocated. Their main drawback comes from their sparse nature, and the lack of detailed geometric information can results in global localization failures when multiple rooms look alike. Another concern is the possible mismatch between the floor plans and the constructed building~\cite{boniardi2017iros}. Li \etalcite{li2020iros} address the scale difference between constructed structure and floor plans by introducing a new state variable. Boniardi \etalcite{boniardi2019iros} uses cameras to infer the room layout via edge extraction and match it against the floor plan. In the evaluation, the authors initialized the pose within 10\,cm and 15$^\circ$ from the ground-truth pose, and did not evaluate global localization. We speculate that edge extraction of the walls is not sufficient in a highly repetitive indoor environment where many rooms have the same size. Both approaches provide tracking capabilities, but not global localization.  

Recent works in extracting semantic information with deep learning models showed significant improvement in performance for both text spotting~\cite{liao2019arxiv}\cite{shi2015arxiv} and object detection~\cite{bochkovskiy2020arxiv}\cite{zhang2022arxiv}. 
The use of textual cues for localization is surprisingly uncommon, with notable works by Cui \etalcite{cui2021iros} and Zimmerman \etalcite{zimmerman2022iros}. Both works considered using textual information within an MCL framework, but used different approaches to integrate it. In our approach, we expand our previous work~\cite{zimmerman2022iros} to consider semantic cues via object detection, not only textual ones.

The use of semantic information for localization and place recognition is applied to a variety of sensors, including 2D and 3D LiDARs, RGB and RGB-D cameras. 
Rottmann \etalcite{rottmann2005aaai} use AdaBoost features from 2D LiDAR scans to infer semantic labels such as office, corridor and kitchen. They combine the semantic information with occupancy grid map in an MCL framework. Unlike our approach, their method requires a detailed map and manually assigning a semantic label to every grid cell. Hendrikx \etalcite{hendrikx2021icra} utilize available building information model to extract both geometric and semantic information, and localize by matching 2D LiDAR-based features corresponding to walls, corners and columns. While the automatic extraction of semantic and geometric maps from a BIM is promising, the approach is not suitable for global localization as it cannot overcome the challenges of a repetitively-structured environment.

Atanasov \etalcite{atanasov2015ijrr} treat semantic objects as landmarks that include their 3D pose, semantic label and possible shape priors. They detect objects using a deformable part model~\cite{felzenszwalb2013acm}, and use their semantic observation model in an MCL framework. The results they report do not outperform LiDAR-based localization. An alternative representation for semantic information is a constellation model, as suggested by Ranganathan \etalcite{ranganathan2007rss}. In their approach, they use stereo cameras, exploiting depth information. They rely on hand-crafted features including SIFT~\cite{lowe1999iccv} to detect objects. Places are associated with constellations of objects, where every object has shape and appearance distribution and a relative transformation to the base location. Unlike these two approaches, our approach does not require exact poses for the semantic objects.
A more flexible representation is proposed by Yi \etalcite{yi2009iros}, who use topological-semantic graphs to represent the environment. They extract topological nodes from an occupancy grid map, and characterize each node by the semantic objects in its vicinity. It suffers when objects are far from the camera and can easily diverge when objects cannot be detected, while our approach is more robust as it relied additionally on LiDAR observations and textual cues. Similarly to the above mentioned approaches, we also use sparse representation for semantic objects. However, by using deep learning to detect objects, we are able to detect a larger variety of objects with greater confidence, and localize in previously unseen places.  

S\"underhauf \etalcite{sunderhauf2016icra} construct semantic maps from camera by assigning a place category to each occupancy grid cell. They use the Places205 ConvNet~\cite{zhou2014nips} to recognize places, and rely on a LiDAR-based SLAM for building the occupancy grid map. The limitation of their approach is in the high level of semantic abstraction. As their work relied on coarse room categorization, it might not be sufficient for global localization in highly repetitive environments.


\section{Our Approach}
\label{sec:main}  
 
\begin{figure}[t]
  \centering
  \includegraphics[width=1.0\linewidth]{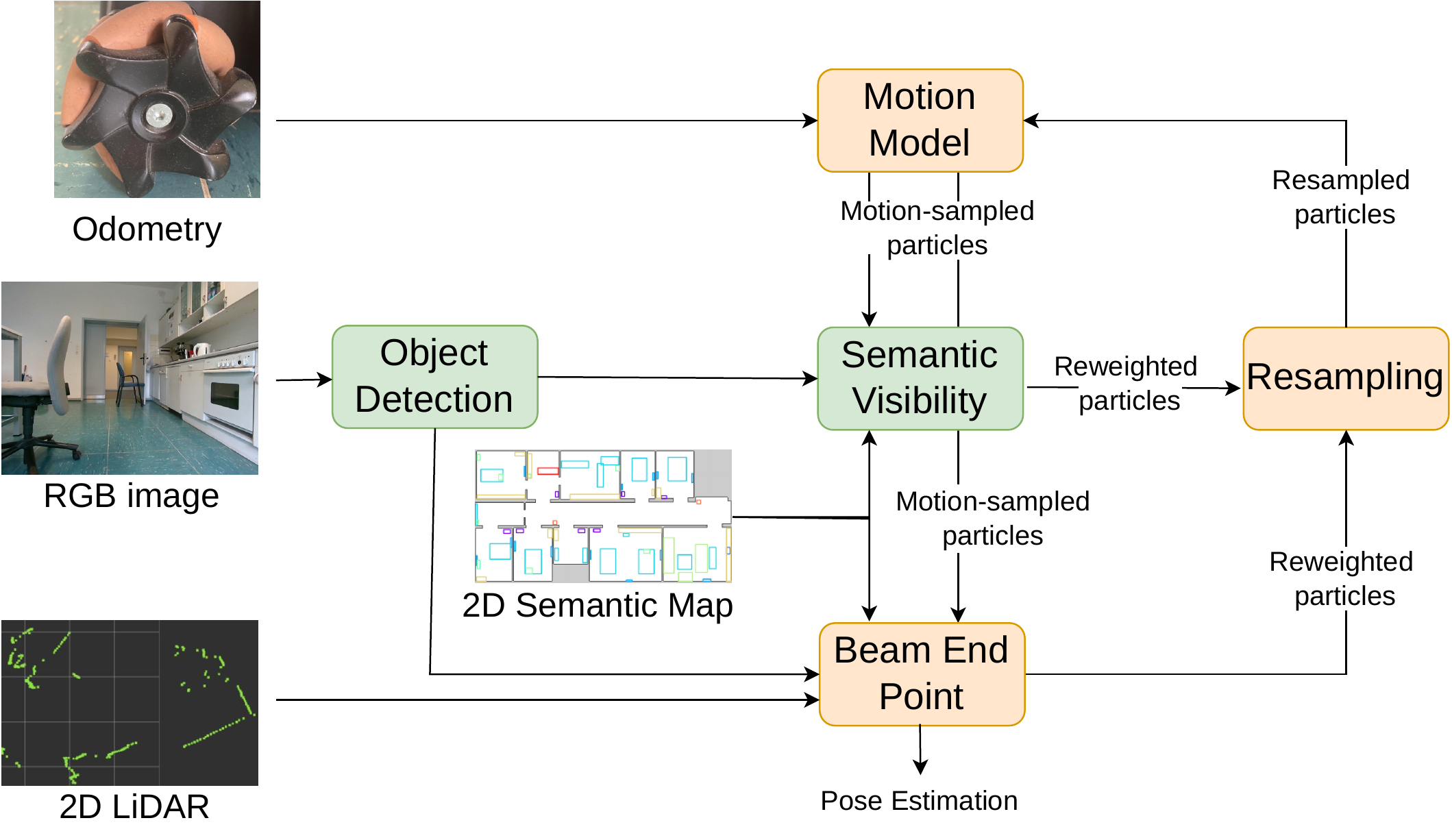}
  \caption{A simplified overivew of the online localization approach. Given RGB images, 2D LiDAR scans an odoemtry input, we integrate semantic cues into an MCL framework.}
  \label{fig:mclflow}
\end{figure}

 Our goal is to globally localize in an indoor environment represented by a nondescript floor plan and a high level semantic map. As sensors for localization, we use 2D LiDAR, cameras and wheel odometry. We build our localization approach on the Monte Carlo localization~(MCL) framework~\cite{dellaert1999icra}. To distinguish between locations that appear similar or identical in the sparse maps, we introduce imprecise, high-level semantic maps in \secref{sec:semmaps} and a sensor model for semantic similarity in \secref{sec:visibility}. The integration of the semantic information in the MCL framework is introduced in \secref{sec:integration}. In addition, we perform an analysis to determine the stability of semantic classes as discussed in \secref{sec:semhierarchy} and utilize the semantic information to discard LiDAR measurements resulting from dynamic objects. Furthermore, in \secref{sec:semhierarchy} we explore a hierarchical semantic approach for inferring the room type based on objection detection, and exploit this information to initialize the particle filter. An overview of the approach is illustrated in \figref{fig:mclflow}.

  \subsection{Monte Carlo Localization}\label{sec:MCL}

Monte Carlo localization \cite{dellaert1999icra} is a particle filter-based approach for state estimation given a map $m$ and sensor readings $z_t$ at time $t$. As we localize in floor plan maps, the robot's state $x_t$ is defined by the 2D coordinates $(x,y)^\top$ and the orientation $\theta \in [0,2\pi)$. The map $m$ is represented by an occupancy grid map~\cite{moravec1989sdsr} or an abstract semantic map, see \secref{sec:semmaps}, and the observation ${z}$ is composed of $K$ elements $z_k$. We apply a recursive Bayesian update to a set of particles $\mathcal{S}_t$, which represent the belief about the robot's pose, $p(x_t\mid z_{1:t}, m)$. Every particle is represented by a state $x_t^{(i)}$ and a weight $w_t^{(i)}$. The proposal distribution $p(x_t \mid x_{t-1}, u_t)$ is sampled when a new motion prior $u_t$ is available, using a motion model for  holonomic robots with odometry noise $\sigma_{\text{odom}} \in \mathbb{R}^3$. By computing the likelihood of an observation ${z_t}$ given a robot's state $x_t$ using the observation model $p({z_t}\mid x_t^{(i)}, m)$ , an individual importance weight  $w_t^{(i)}$ is assigned to each particle. In the resampling step, we use low-variance resampling~\cite{thrun2005probrobbook}.



\subsection{High-Level Semantic Maps} \label{sec:semmaps}
\begin{figure}[t]
  \centering
  \includegraphics[width=0.75\linewidth]{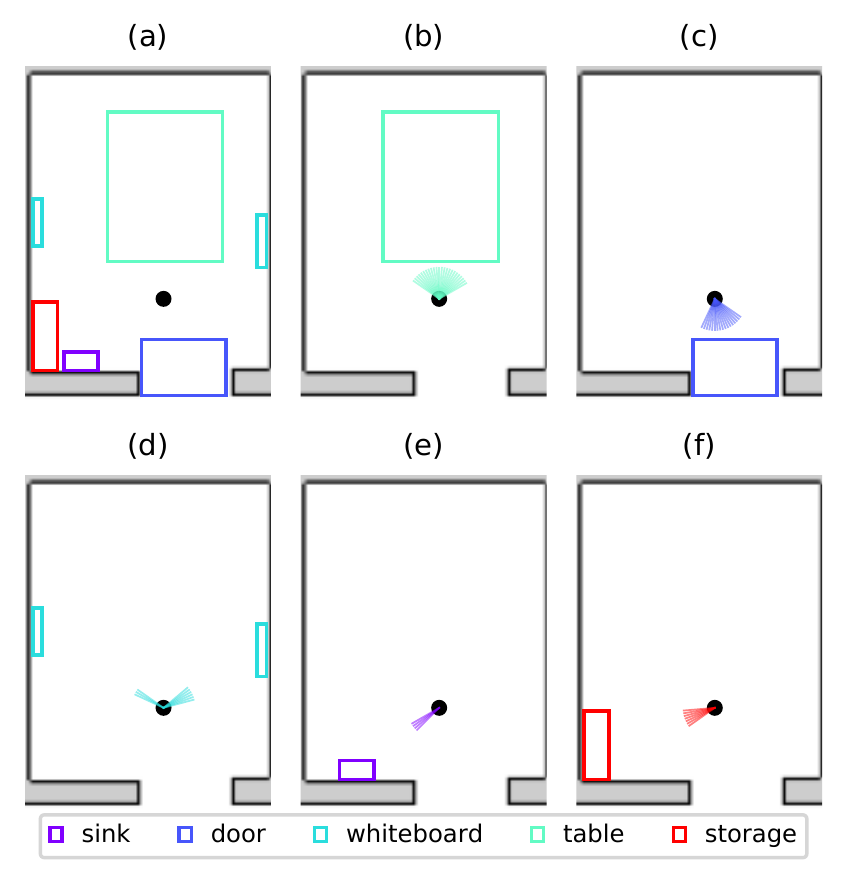}
  \caption{A visualization of the semantic visibility concept. (a) A semantic map of a single room, with a query point (black dot). (b)-(f) The bearings in which each semantic class objects are visible from the query point.}
  \label{fig:semanticvisibility}
\end{figure}

We represent our prior information about semantics with a 2D high level semantic map, where semantic objects are represented by a semantic class label $l$ and a rectangle overlying the occupancy grid map. see \figref{fig:semanticvisibility}. The size of the rectangle does not have to be very accurate and the location where it is marked can be a rough estimate of its actual placement. In our abstract map, objects differed from their actual size by 62.5\%, or up to 1.25\,m. This imprecise representation of semantic information is both generic enough to address variety of objects and simple enough to allow editing by end users. Each room can be assigned a name, corresponding to a text sign, and a room category representing a higher level of semantic understanding compared to basic object detection. The semantic maps can be easily created and edited using the GUI application MAPhisto\footnote{https://github.com/FullMetalNicky/Maphisto}. 
 
\subsection{Beam End Point Model}\label{sec:beamend}
The beam end point model~\cite{thrun2005probrobbook} $p_L(z_t\mid x_t,m)$ is an observation model for range sensors 
 
\begin{align}
 p_L(z_t^k\mid x_t,m_L) &=  \frac{1}{\sqrt{2 \pi \sigma_{\text{obs}}}} \exp{\left(-\frac{\mathrm{edt}(\hat{z}_t^k)^2}{2 \sigma^2}\right)}
 \label{eq:beam_end_model}
\end{align}
Where $\hat{z}_t^k$ is the end point of the LiDAR beam in the occupancy grid map~$m_L$ and edt is the Euclidean distance transform~\cite{felzenszwalb2012toc}, in which each cell is labeled with the distance to an occupied cell in the occupancy grid map. The edt was truncated at $r_{\text{max}}$, a predefined maximal range.


\subsection{Semantic Visibility Model} \label{sec:visibility}

 The last decade's progress in semantic interpretation allows us to use deep learning models for text spotting~\cite{liao2019arxiv}\cite{shi2015arxiv} and object detection~\cite{bochkovskiy2020arxiv}\cite{zhang2022arxiv}. 

\begin{figure}[t]
  \centering
  \includegraphics[width=0.95\linewidth]{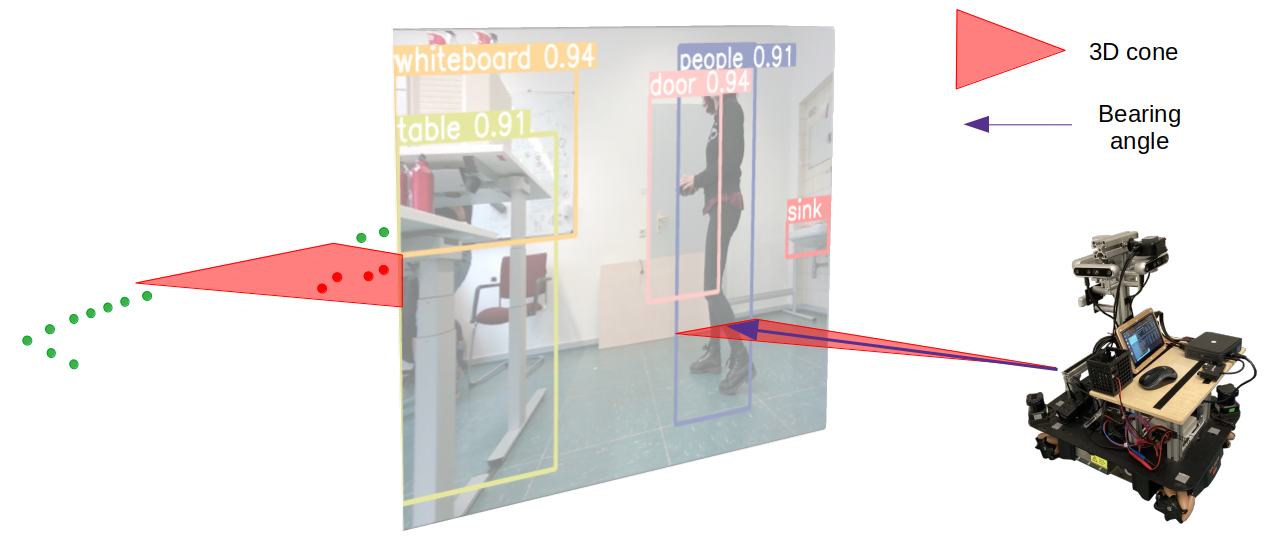}
  \caption{The bounding box detecting a dynamic class (person) is projected to 3D and used to mask the LiDAR beams that fall within the cone.}
  \label{fig:lidarblock}
\end{figure}

  
Object detection is the task of detecting instances of semantic objects in images and videos. In our approach, the required output from an object detection model is a semantic label, a bounding box and a confidence score for every detected object. For each bounding box in the prediction, we transform it to a 3D cone $\boldsymbol{\hat{x}}$ in the robot coordinate system, see \figref{fig:lidarblock}. We take the pixel coordinates of the right and left boundaries of a bounding box, $\{bb_r, bb_l\}$ and project them to 3D rays by using the camera's intrinsics and extrinsics matrices. For a pixel $v=(x, y, 1)^T$, we define the associated 3D ray $V(\lambda)$ as follows:
 \begin{align}
 V(\lambda) = O + \lambda R^{-1}K^{-1}v,
\end{align}
where $K \in \mathbb{R}^{3 \times 3}$ is the camera intrinsics,  $R \in \mathbb{R}^{3 \times 3}$ is the camera rotation and $O\in \mathbb{R}^3$ is camera center.

From the high-level semantic information, we construct visibility maps for the semantic classes. For each valid, free space cell $c$ in the occupancy grid map, we compute the visibility of semantic objects. A semantic object $o$ is visible from a grid cell $c$ if we can ray-trace it without crossing a non-valid, i.e., occupied or unknown, cell. For each cell $c$, we maintain a list of all visible semantic classes. For each semantic class $l$, we store the set of bearing vectors, $\mathcal{B}=\{\boldsymbol{b}_1,\dots,\boldsymbol{b}_n\}, \norm{\boldsymbol{b}_i}=1$, in which objects of class $l$ are visible. This process of constructing the visibility maps is performed once, when the algorithm is launched, and is illustrated in \figref{fig:semanticvisibility}.

A semantic observation $y_t$ includes the set of detected objects. For every object we store its semantic label $l$, its confidence score $f$ and the center of its cone as the bearing~$\boldsymbol{\hat{b}}$. For each particle $s_t^{(i)}$ with pose $x_t^{(i)}=(x, y, \theta)^\top$, we transform the bearing $\boldsymbol{\hat{b}}$ into the world coordinate system. We query the pre-built semantic visibility maps for cell $c$ corresponding to the pose of particle $s_t^{(i)}$, and compare it with the observation. If an object is observed with confidence score $f$ which is lower than a threshold $\tau$, we ignore the observation. Otherwise, if an object with semantic label $l$ is visible from cell $c$, we compare the observed bearing $\boldsymbol{\hat{b}}$ to the set of possible bearings $\mathcal{B}$ by using the cosine similarity:
 \begin{align}
  \text{sim}(\boldsymbol{v}_1, \boldsymbol{v}_2) = \frac{\boldsymbol{v}_1 \cdot \boldsymbol{v}_2}{\norm{\boldsymbol{v}_1} \norm{\boldsymbol{v}_2}}.
\end{align}
where $\boldsymbol{v}_1, \boldsymbol{v}_2 \in \mathbb{R}^2$.
\indent To compare the observed bearing $\boldsymbol{\hat{b}}$ with all possible visible bearings ${\boldsymbol{b}_i} \in \mathcal{B}$ and select the best match according to the distance~$d$, defined as
\begin{align}
  d = 1 - \max_{\boldsymbol{b}_i \in \mathcal{B}} (\text{sim}(\boldsymbol{b}_i, \boldsymbol{\hat{b}})).
   \label{eq:semantic_distance}
\end{align}
\indent For a set of detected objects ${z_t^S}$, our observation model is given by 
\begin{align}
 p_S(z_t^k\mid x_t,m_S) &=  \exp{\left(-d \right)} \hspace{0.5cm} z_t^k \in z_t^S,
 \label{eq:sem_viz_model}
\end{align} 
where $z_t^k$ is the $k\textsuperscript{th}$ confidently observed object in the set ${z_t^S}$, and $m_S$ is the abstract semantic map.

\subsection{Integrating Different Modalities in the MCL Framework} \label{sec:integration}

We handle all information sources asynchronously -- the motion model is sampled when odometry input is available, and the particles are re-weighted when an observation arrives. We integrate the 2D LiDAR measurements and the object detections using two different observation models.
For a 2D LiDAR observation, ${z_t^L}$, we use the beam-end model $p_L({z_t^L}\mid x_t,m_L)$ described in \secref{sec:beamend}. When object detection information arrives, ${z_t^S}$, we use the semantic visibility model $p_S({z_t^S}\mid x_t,m_S)$, detailed in \secref{sec:semstability}.

The product of likelihood model assumes elements of each observation, e.g scan points in a LiDAR scan, are independent of each other. With the high angular resolution of our LiDAR this assumption does not hold. Similarly, for the semantic visibility model, detected objects are not entirely independent of each other as they often belong to the same context. The traditional product of likelihood model tends to be overconfident in such circumstances, leading us to choose the product of experts model \cite{miyagusuku2019ral}, which uses geometric mean to compute the weight of each particle  
\begin{align}
 p({z_t}\mid x_t,m) &= \prod_{k=0}^K p(z_t^k\mid x_t,m)^{\frac{1}{K}},
\end{align}
where $z_t^k$ is a single component of an observation ${z_t}$, be it a LiDAR scan or a set of detected objects. The beam-end model is triggered only when the robot moves more than $d_{xy}$ or rotates more than $d_{\theta}$, while the semantic observation model is always updated. Based on the semantic stability analysis, we detected semantic classes that tend to move frequently, which we refer to as dynamics. In addition to excluding these classes from the semantic map, we also use these detections to filter out LiDAR measurements that are the result of dynamics, as seen in \figref{fig:lidarblock}

 \subsection{Semantic Stability Analysis} \label{sec:semstability}
 To decide which semantic classes would benefit localization, we estimated how likely they are to move around. We prepared a semantic map for all detectable classes, and examined the training-dedicated recordings T1-T5 spanning over multiple weeks. As our dataset includes the ground truth pose of the robot using an external reference system, we were able to conclude whether the position of detected objects corresponded to their position in the map. Using \eqref{eq:semantic_distance}, we consider a detected object to correspond to the semantic map if $d < \tau_{s}$.  We calculate the ratio of map-consistent detection per semantic class, and deem a semantic class stable if the ratio was above 0.6. The ratio is computed by dividing the number of map-consistent detected objects of class l, by the total number of detections. Unstable classes are excluded from the semantic visibility model, and then stability scores are given in \tabref{tab:stab_score}.

  \begin{table*}[t] 
  \caption{Semantic stability scores for different detected object classes computed on sequences T1-T5.}
  \centering 
  \resizebox{\textwidth}{!}{
\begin{tabular}{ccccccccccccccc}\toprule

Class        & sink & door & oven & whiteboard & table & cardboard & plant & drawers & sofa & storage & chair & extinguisher & person & desk\\ \midrule
Score        & 0.97 & 0.96 & 0.90 & 0.91 & 0.95 & 0.46 & 0.88 & 0.86 & 0.99 & 0.96 & 0.58 & 0.84 & 0.11 & 1.00\\
\bottomrule 
\end{tabular}
}
\label{tab:stab_score} 
\end{table*}

\subsection{Hierarchical Semantic Localization} \label{sec:semhierarchy}
In big indoor environments, a very large number of particles is required to sufficiently cover the area in the initialization phase of global localization, which result in great computational costs. It is possible to reduce the number of used particles and achieve global localization by considering a hierarchy of semantic information. We propose to infer the room category (office, corridor, kitchen, reception) based on the predictions from the object detection. We use a nearest-neighbor classifier~\cite{mucherino2009springer} to learn a relationship between the detected objects and the room category. We encode the semantic information as a feature vector $r \in \mathbb{R}^M$, where $M$ is the number of classes we are able to detect. Each vector element $r_l$ represents the number of detected objects from a specific semantic label $l$. We used our initial semantic observations to infer the room category, and initialize the particle filter accordingly, so that particles are only initialized in rooms of the same category. The information about the category of each room is stored in the high-level semantic map~(\secref{sec:semmaps}).

\section{Experimental Evaluation}
\label{sec:exp}

%

The focus of this work is to provide an efficient, robust localization approach that exploits semantic information for long-term operation in sparse floor plans. 
%
We conducted our experiments to support our claims and show that our approach is able to: 
(i) localize in sparse floor plan-like map with high symmetry using semantic cues, 
(ii) localize long-term without updating the map,
(iii) localize in previously unseen environment,
(iv) localize the robot online using an onboard computer.

\begin{table}[t]
  \caption{Algorithm parameters}
   \centering
   \resizebox{\columnwidth}{!}{
 \begin{tabular}{cccccccc}\toprule
 Method & $\sigma_{\text{odom}}$  &  $\sigma_{\text{obs}}$ & $r_{\text{max}}$ & $ \tau_{s} $ & $\rho$ & $d_{\text{xy}}$ & $d_{\theta}$\\ \midrule
 MCL & (0.15\,m, 0.15\,m, 0.15\,rad) & 6.0 & 15.0\,m & - & - & 0.1\,m & 0.03\,rad\\
 TMCL & (0.15\,m, 0.15\,m, 0.15\,rad) & 6.0 & 15.0\,m & - & 0.5 & 0.1\,m & 0.03\,rad\\
 HSMCL & (0.15\,m, 0.15\,m, 0.15\,rad) & 6.0 & 15.0\,m & 0.6 & - & 0.1\,m & 0.03\,rad\\

  \bottomrule
 \end{tabular}
 }
 \label{tab:parameters}
\end{table}

\subsection{Experimental Setup}

\begin{figure}[t]
  \centering
  \includegraphics[width=0.80\linewidth]{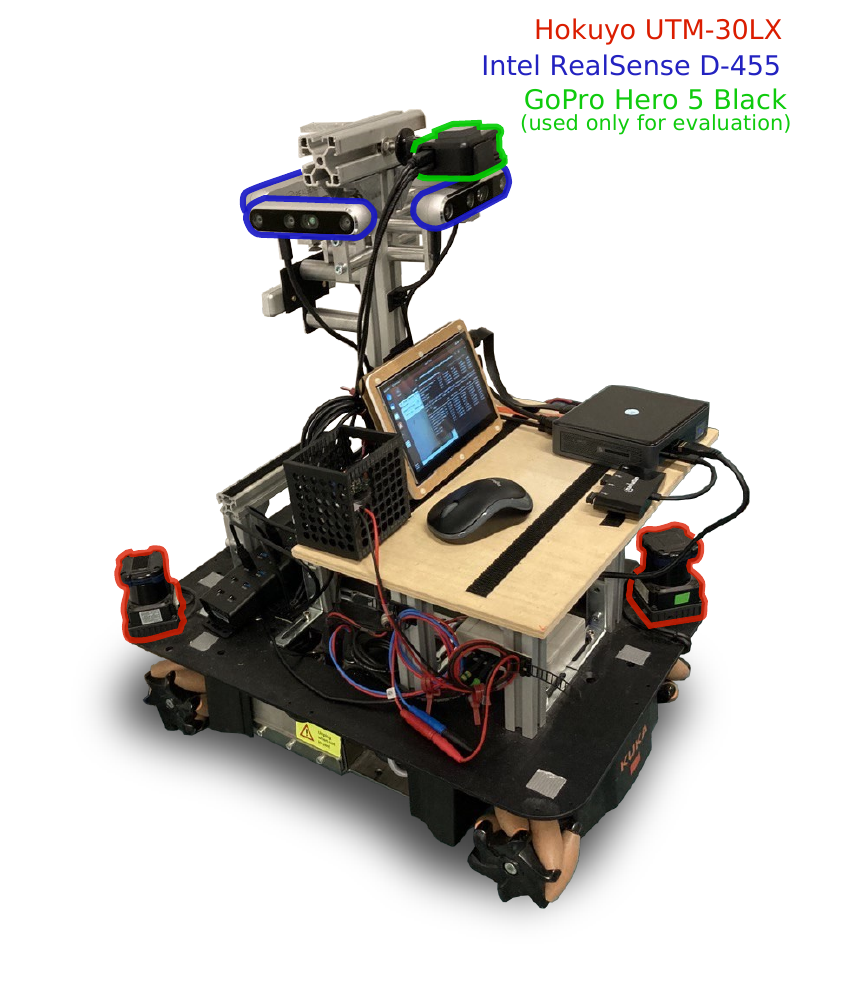}
  \vspace{-3mm}
  \caption{The data collection platform, an omnidirectional Kuka YouBot, with 2D LiDAR scanners (marked by a red outline) and with 4 cameras (marked by a blue outline) providing $360^{\circ}$ coverage. The up-ward facing camera (marked by a green outline) is only used for generating the ground truth via AprilTag detections.}
  \label{fig:robot_sensors}
\end{figure}

To evaluate the performance of our approach, we recorded a dataset on the first and second floors of our building. Our mobile sensing platform consisted of a Kuka YouBot platform with 2 Hokuyo UTM-30LX LiDAR sensors, wheel encoders, 4 cameras covering jointly a $360^{\circ}$ field-of-view, and an upward-looking camera that is used only for evaluation purposes, see\figref{fig:robot_sensors}. The recordings span across several weeks, capturing different scenarios including moving furniture, opening and closing of doors and humans passing by.  


By using precisely localized AprilTags \cite{olson2011icra-aara}, which are densely placed (approx.~$\text{1 tag/m}^2$) on the ceiling of every room and corridor on the second floor, we are able to extract the ground truth pose of the robot from the upwards-looking camera. The camera is used to detect the AprilTags, which allows us to accurately localize the robot even when the environment undergoes changes. The upward-looking camera captured frames at 25\,fps, and due to its wide-angle lens, we were able to detect multiple AprilTags in every frame. The pose was extracted in a least-squares fashion using multiple detections. The locations of the AprilTags were obtained using a high resolution terrestrial FARO laser scanner, and were aligned to the floor plan of the second floor. By enforcing a shared coordinate system, we are able to compare the pose estimation to our ground truth poses. 

  Recording R1-R11 are captured in the second floor of our building and include ground truth poses. Recording Q1-Q3 were recorded in the first floor of the building and do not include ground truth information, and are used for qualitatively evaluation on previously unseen environment. Each sequence was evaluated multiple times to account for the inherent stochasticity of the MCL framework. 

In our implementation, we used YOLOv5~\cite{jocher2020zenodo}, which is a family of object detection models of varying size and performance. YOLOv5 models are capable of real-time inference on CPU-only platforms, thus making them well-suited for mobile robots. We trained a small model, YOLOv5s, on 581 images from the second floor of our building using the default training script provided in the \href{https://github.com/ultralytics/yolov5}{YOLOv5 repository}. The map used for localization is joint map of two CAD floor plan drawings, of the first and second floor side-by-side, illustrated in \figref{fig:motivation2}. The semantic information was integration using our GUI application MAPhisto\footnote{https://github.com/FullMetalNicky/Maphisto} based on our recollection of location of semantic objects. For all experiments, we use a map resolution of 0.05\,m by 0.05\,m per cell and the algorithm parameters specified in \tabref{tab:parameters}. 

  As baseline, we compare against AMCL~\cite{pfaff2006springerstaradvanced}, which is a publicly available and highly-used ROS package for MCL-based localization,  our own MCL implementation without using semantic cues and a text-enhanced MCL~\cite{zimmerman2022iros}, which we refer to as TMCL. Our method, exploiting both semantic information from object detection and hierarchical semantic knowledge discussed in \secref{sec:semhierarchy}, is referred to as HSMCL. For the tracking experiments, we considered SMCL, a variation of our approach that uses only semantic cues through the semantic visibility model, without hierarchical semantic localization. 
All experiments were executed with 10,000 particles in the filter unless mentioned otherwise.

\begin{table}[t] 
  \caption{Success rate for 11 sequences recorded all across the second floor in the span of several weeks. A run was considered successful if the algorithm converged to the ground truth in the first 95\% of the recording and remained localized until the end of the sequence. }
  \centering 
 
\begin{tabular}{ccccccc}
\toprule

Method        & R1 & R2 & R3 & R4 & R5 & R6 \\ \midrule
AMCL  & 0\%   & 0\%   & 0\%   & 0\%   & 0\%   & 0\%   \\\
MCL & 40\%  & 20\%  & 60\%  & 40\%  & 20\%  & 0\%  \\
TMCL  & 80\%  & 0\%   & 100\%   & 80\%  & 60\%  & 40\%  \\
HSMCL & 100\%   & 100\%   & 100\%   & 100\%   & 100\%   & 100\%  \\

\bottomrule 
\toprule

Method        &  R7 & R8 & R9 & R10 & R11 & AVG\\ \midrule
AMCL    & 0\%   & 0\%   & 0\%   & 0\%   & 0\%   & 0\% \\
MCL   & 0\%   & 40\%  & 20\%  & 60\%  & 0\%   & 27\% \\
TMCL   & 100\%   & 100\%   & 100\%   & 80\%  & 100\%   & 76\% \\
HSMCL    & 100\%   & 100\%   & 100\%   & 100\%   & 100\%   & 100\% \\

\bottomrule 
\end{tabular}

\label{tab:success_r1r10} 
\end{table}

We consider two metrics, the success rate and absolute trajectory error~(ATE) after convergence. In our definition, convergence occurs when the estimated position is within a radius of  0.3\,m from the ground truth pose and the estimated orientation is within $\frac{\pi}{4}$\,radians. Tracking of the pose is considered unreliable if the pose estimate diverges for more than 1\% of the time. If convergence did not occur within the first 95\% of the sequence, or if the pose is not reliably tracked  from convergence moment until the end of the sequence, we consider it a failure.



\begin{table*}[t] 
  \caption{ATE for global localization on consistently stable (100\% success rate) sequences recorded all across the second floor in the span of several weeks. Angular error
in radians~/~translational error in meters.}
  \centering 
  \resizebox{\textwidth}{!}{
    \setlength\tabcolsep{4.5pt}
\begin{tabular}{ccccccccccccc}\toprule

Method        & R1 & R2 & R3 & R4 & R5 & R6 & R7 & R8 & R9 & R10 & R11 & AVG\\ \midrule
AMCL  & {-}/{-}     & {-}/{-}     & {-}/{-}     & {-}/{-}     & {-}/{-}     & {-}/{-}     & {-}/{-}     & {-}/{-}     & {-}/{-}     & {-}/{-}     & {-}/{-}     & {-}/{-}\\
MCL & {-}/{-}     & {-}/{-}     & {-}/{-}     & {-}/{-}     & {-}/{-}     & {-}/{-}     & {-}/{-}     & {-}/{-}     & {-}/{-}     & {-}/{-}     & {-}/{-}     & {-}/{-}\\
TMCL  & {-}/{-}     & {-}/{-}     & 0.048/0.16    & {-}/{-}     & {-}/{-}     & {-}/{-}     & 0.034/0.22    & 0.043/0.18    & 0.050/0.21    & {-}/{-}     & 0.034/0.18    & 0.042/0.19 \\
HSMCL & 0.054/0.15    & 0.064/0.24    & 0.069/0.25    & 0.205/0.23    & 0.100/0.34    & 0.064/0.23    & 0.069/0.23    & 0.049/0.18  & 0.090/0.26    & 0.052/0.16    & 0.052/0.25    & 0.079/0.23 \\
\bottomrule 
\end{tabular}
}
\label{tab:ate_r1r10} 
\end{table*}

\subsection{Long-Term Localization in CAD Floor Plans}

The first experiment evaluates the performance of our approach and supports the claim that we are capable of long-term localization in sparse, floor-plan-like maps. Sequences R1-R11 are recorded in April-June 2022, and traverse all the rooms in the second floor. The given map had been constructed in 2021. All sequences include humans walking around, opening and closing of doors, moving furniture and large amount of clutter. We repeat the evaluation of each sequence 5 times, computing the success rate, ATE and convergence time over all 5 runs, and compare against the baselines. As can be seen in \tabref{tab:success_r1r10} the semantically-enhanced methods have superior performance over the baselines. AMCL and MCL are mostly used with detailed maps constructed using range-sensor measurements, and we can attribute their poor performance to the sparse nature of the floor plans. This highlights the impact of semantic information when localizing in nondescript, sparse maps, especially in face of high geometric symmetry. 

As reported in \tabref{tab:ate_r1r10}, upon successful convergence, HSMCL achieves accuracy of 0.23\,m and negligible angular error. HSMCL successfully converges, on average over all sequences, after 25\,s. 

We further provide pose tracking experiments. A similar approach to ours, Boniardi \etalcite{boniardi2019iros}, tracked the pose of a robot by inferring the room layout from camera images, reporting RMSE of approx.~0.23\,m and approx.~0.04\,rad with adaptive particle number ranging between 1,500-5,000. However, they did not provide open source code. Our office environment is similar to the Freiburg one where Boniardi \etalcite{boniardi2019iros} evaluated their method. For the tracking experiments we used SMCL, which integrates semantic cues from object detection, without hierarchical information. We report our tracking results with fixed 1500 particles in \tabref{tab:tracking_ate}, achieving an ATE of 0.2\,m and 0.05\,rad. This suggest that integrating semantic cues, and specifically, our SMCL approach, are beneficial also for tracking purposes and not only for global localization.

\begin{table}[t] 
  \caption{ATE for tracking on a subset of sequences recorded all across the second floor in the span of several weeks. The particle filter was set to adaptive 1,500-5,000 particles for AMCL and a fixed 1,500 particles for MCL and SMCL. Angular error
in radians / translational error in meters.} 
  \centering 
  \resizebox{\columnwidth}{!}{
    \setlength\tabcolsep{4.5pt}
\begin{tabular}{cccccccc}\toprule

Method        &  R3 & R4 &  R6 & R7 &  R8  & R10 & AVG\\ \midrule
AMCL  & {-}/{-}     & {-}/{-}     & 0.047/0.22    & {-}/{-}     & {-}/{-}     & {-}/{-}     & 0.047/0.22 \\
MCL & 0.051/0.17    & 0.050/0.21    & 0.051/0.29    & 0.064/0.23    & 0.039/0.14    & 0.041/0.15    & 0.049/0.20 \\
SMCL  & 0.063/0.21    & 0.046/0.22    & 0.068/0.29    & 0.048/0.19    & 0.042/0.15    & 0.044/0.13    & 0.052/0.20 \\
 \bottomrule 
\end{tabular}
}
\label{tab:tracking_ate} 
\end{table}

\subsection{Localization in a Previously Unseen Environment}

\begin{figure}[t]
  \centering
  \includegraphics[width=0.95\linewidth]{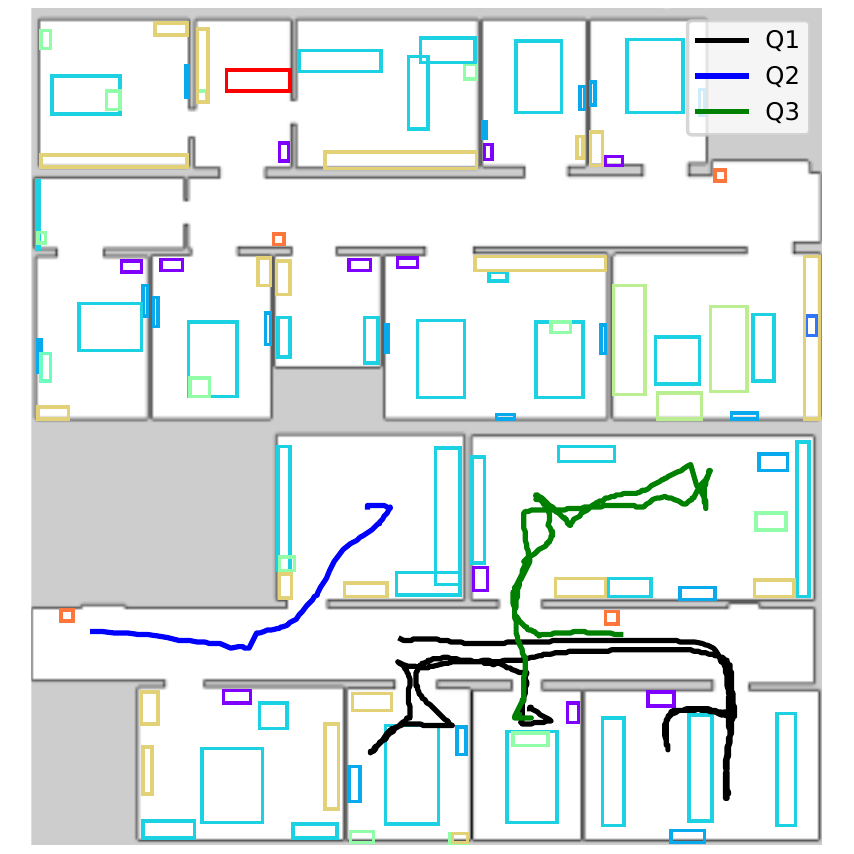}
  \caption{Examples of pose estimation for localization in previous unseen environment, using SMCL and 10,000 particles.}
  \label{fig:q1q2}
\end{figure}

To support our claim that we are able to localize in a previously unseen environment, we qualitatively evaluate our method on sequences Q1-Q3 recorded on the first floor of our building. The object detection model and the room category classifier were not trained or validated on data from this floor. While the first floor is not entirely dissimilar to the second one, it does include different furniture and rooms that serve different purposes such as a classroom and a robotics lab. The pose estimated by SMCL sequences is shown in \figref{fig:q1q2}. Our approach correctly predicts that the robot is located in the first floor and identifies the correct room and maintaining a trajectory that is consistent with floor plan map.

  
\subsection{Ablation Study}
To justify our use of both low-level and hierarchical semantic information, we conducted an ablation study. We analyzed three strategies for integrating semantic knowledge into an MCL framework. SMCL uses only semantic cues through the semantic visibility model. HMCL uses semantic hierarchy, described in \secref{sec:semhierarchy}, to initialize the particles only in the rooms corresponding to the observed room category, and then relies solely on the LiDAR information. HSMCL combines both strategies. The ATE was computed only on stable sequences with 100\% success rate. As can be seen in \tabref{tab:ablation}, utilizing the two levels of semantic information benefits localization. HSMCL was able to localize stably even on the challenging sequences, where other methods failed. The ATE for HSMCL is on par with the other methods, and the slightly larger error can be attributed to including more challenging sequences and runs in the computation of the ATE for HSMCL, sequences and runs where other methods failed to localize entirely.

\begin{table}[t] 
  \caption{Performance on 11 sequences recorded all across the second floor in the span of several weeks. A run was considered successful if the algorithm converged to the ground truth in the first 95\% of the recording and remained localized until the end of the sequence. Angular error in radians / translational error in meters.}
  \centering 
  \resizebox{\columnwidth}{!}{
\begin{tabular}{cccccc}\toprule

Method & Hierarchy & Semantics & Success & ATE  & ATE \\
 & & & & (\# of stable sequences) & (\# of successful runs) \\  \midrule
MCL & & & 27\% & -  (0) & 0.046/0.20 (15)\\
HMCL & \checkmark &  & 61\% & 0.046/0.21 (3) &  0.044/0.19 (34)\\
SMCL &  & \checkmark &   81\% &  0.055/0.23 (7) & 0.066/0.24 (45)\\
HSMCL & \checkmark & \checkmark  & 100\% & 0.079/0.23 (11) & 0.079/0.23 (55)\\
\bottomrule 
\end{tabular}
}
\label{tab:ablation} 
\end{table}

\subsection{Runtime}


We evaluate the runtime performance of our approach in support of our fourth claim, that we are able to operate onboard and allow real-time localization. We tested our approach on a Dell Precision-3640-Tower (with NVidia GeForce RTX 2080) and once on an Intel NUC10i7FNK, which we have on our robot. 
The Dell PC has  64\,GB of RAM and runs at 3.70\,GHz. The Intel NUC has 16\,GB of RAM and runs at 1.10\,GHz.
 The measurements are reported in \tabref{tab:performance}. Since we are using 4 cameras simultaneously for object detection, we used an optimized ONNX export of YOLOv5s, and run inference on 320 by 240 images. Qualitative online tests indicates that reducing the resolution does not impact the detection accuracy significantly. These runtime results suggest that our approach is suitable for online localization, and utilizes semantic information without requiring a GPU onboard.

\begin{table}[t] 
  \caption{Runtime for HSMCL, with 10,000 particles. The Yolov5s results are for inference on a single camera.}
  \centering 
  \resizebox{\columnwidth}{!}{
\begin{tabular}{ccccc}\toprule

Platform        &  Sem. Visibility & Beam-End & Yolov5s & Yolov5s\\ 

 & & & (640x480) & (320x240)\\ \midrule
NUC10i7FNK  & 55\,ms & 24\,ms & 223\,ms& 57\,ms  \\ 
Dell Precision-3640-Tower & 19\,ms & 14\,ms & 10\,ms & 6.8\,ms\\

\bottomrule 
\end{tabular}
}
\label{tab:performance} 
\end{table}








\section{Conclusion}
\label{sec:conclusion}

Our approach incorporates semantic information, from low-level object detection to higher understanding of room categories, to assist navigation in human-oriented environments. This enables us to successfully localize in sparse floor plans under high geometric symmetry and changing environments. We demonstrate that using sparse and abstract map representation benefits long-term localization, and reduces the need to update the map. We also provide a tool for updating the semantic map, when critical changes occur. For our evaluation, we recorded a dataset spanning across weeks, introducing a variety of elements that are not represented in the floor plan, and the changes a human-occupied environment undergoes. We compared our performance to other existing methods, supporting all of our claims. The results of our experiments imply mobile localization systems can benefits greatly from exploiting ever-present semantic cues.


\section*{Acknowledgments}
We thank Lior Rozin for his contribution to the GUI application MAPhsito. We are also grateful to segments.ai for supporting our research with free and full access to all features. We also thank Matteo Sodano, Louis Wiesmann and Thomas L\"abe for their assistance with the data collection.

\bibliographystyle{plain_abbrv}

\bibliography{glorified,new}

\end{document}